\documentclass{article}
\usepackage{amsmath,graphicx,mlspconf,amssymb}
\usepackage{bm}
\usepackage{hyperref}
\hypersetup{
        colorlinks = true,
        allcolors = blue,
        breaklinks = true,
        linktocpage=true,
}
\usepackage{multirow}

\DeclareMathOperator{\Att}{Attention}
\DeclareMathOperator{\softmax}{softmax}
\title{Language-Oriented Semantic Latent Representation \\for Image Transmission}
\name{Giordano Cicchetti$^*$, Eleonora Grassucci$^*$, Jihong Park$^\dagger$, Jinho Choi$^\dagger$, Sergio Barbarossa$^*$, \\
\textit{and Danilo Comminiello}$^*$
\thanks{This work was supported by the European Union under the Italian National Recovery and Resilience Plan (PNRR) of NextGenerationEU, partnership on ``Telecommunications of the Future” (PE00000001 - program RESTART).}}
\address{$^*$Dept. of Information Engineering, Electronics, and Telecomm., Sapienza University of Rome, Italy\\
$^\dagger$School of Information Technology, Deakin University, Australia}

\begin{document}
%\ninept

\maketitle

\begin{abstract}

%In the new paradigm of semantic communication, more focus is given to the exchange of semantic information between sender and receiver, therefore the choice for a proper and meaningful semantic vector is crucial for content regeneration at the receiver side. 
%Different approaches try to encapsulate the semantics of the data in its textual description, exploiting powerful data-to-text models. This information is then used to reconstruct data at the receiver side using text-to-data generative models.
%However, text alone may not be able to completely fulfill the intrinsic meaning of data and although reconstructed data are semantically aligned, they might be perceptually very different from the original one. 
%In this paper, we propose to use textual descriptions along with a compressed embedding representation of data as a semantic carrier. Using also image embeddings to regenerate content allows for more precise content generation and brings superior performance in perceptual metrics. 
%Experimental results for the image-to-image transmission task validate the potential of our approach, which can achieve high perceptual similarities with fewer transmitted bits, despite the adverse conditions due to the noisy communication channels.

In the new paradigm of semantic communication (SC), the focus is on delivering meanings behind bits by extracting semantic information from raw data. Recent advances in data-to-text models facilitate language-oriented SC, particularly for text-transformed image communication via image-to-text (I2T) encoding and text-to-image (T2I) decoding. However, although semantically aligned, the text is too coarse to precisely capture sophisticated visual features such as spatial locations, color, and texture, incurring a significant perceptual difference between intended and reconstructed images. To address this limitation, in this paper, we propose a novel language-oriented SC framework that communicates both text and a compressed image embedding and combines them using a latent diffusion model to reconstruct the intended image. Experimental results validate the potential of our approach, which transmits only 2.09\% of the original image size while achieving higher perceptual similarities in noisy communication channels compared to a baseline SC method that communicates only through text.
The code is available at \href{https://github.com/ispamm/Img2Img-SC/}{https://github.com/ispamm/Img2Img-SC/}.
\end{abstract}

\begin{figure}[!t]
\centering
\includegraphics[width=0.90\linewidth,keepaspectratio]{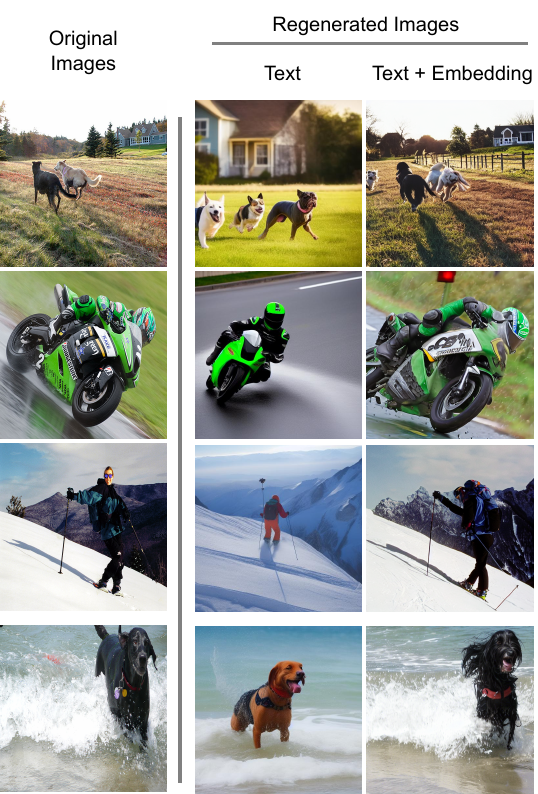}
\caption{Random samples from Flickr-8k dataset \cite{Hodosh_Young_Hockenmaier_2013}. On the left-hand side original images. On the right-hand side regenerated images using our framework and different conditioning signals. Images regenerated using both textual description and image embeddings are not only semantically aligned but also perceptually very similar to the original ones. }
\label{fig_1}
\vspace{-0.3cm}
\end{figure}

\section{Introduction}
\label{sec:intro}
In the field of communication theory and technology, the quest for more meaningful, efficient, and effective exchanges of information has led to the emergence of a fascinating research area known as Semantic Communication (SC). 
The key idea behind SC is to convey the semantic information of data, which may also be attached to a compressed representation of the content, rather than exchanging the whole message. Especially, semantics may be useful if not enough bandwidth is available for the whole data transmission.

% not only data but also information related to their semantics so that the receiver agent can reconstruct the original intended message if something during the transmission goes wrong.

%of the data  

The sender cares to extrapolate and encapsulate the semantics of the data it wants to transmit in a compact and meaningful representation. The receiver uses the received information to reconstruct semantically equivalent data rather than the original content. \cite{Qin2021SemanticCP, Seo2022TowardSC, Gndz2022BeyondTB, Luo2022WC}.
These concepts introduce numerous additional degrees of flexibility that can be strategically leveraged in system design and resource allocation. Nowadays, one of the most appropriate tools that can be employed to solve the newly posed challenges is generative deep learning \cite{Barbarossa2023COMMAG, Han2022GenerativeMB}. 

The recent developments made by deep generative models are well known to everyone. These models can generate almost any kind of multimedia content such as text, images, video, and audio. One of their most interesting features is the ability to generate content starting from semantic conditioning, which can be an extremely compressed version of original data, such as text \cite{Nam2023SECON,nam2024language} or low-dimensional latent vectors \cite{ Nemati2023}. This is an important key point in SC and for the future generation of networks (6G), since the adoption of generative models allows for a substantial reduction in information exchanging, bandwidth requirements, and latency, leaving untouched perceptual results at the receiver side. However, the fidelity of the generation at the receiver strictly depends on the quality of the transmitted semantics. Unfortunately, extracting the proper semantic representation from data is not straightforward and no unique recipes exist to do so. As a matter of fact, the best way to extract meaningful semantics from different types of data is still an open problem \cite{grassucci2024generative}. 

One of the first approaches towards the use of generative models in SC is the employment of a variational autoencoder acting as a transceiver in a Deep joint source and channel coding (DeepJSCC) system \cite{Gndz2022BeyondTB}. The variational autoencoder (VAE) adopted performs perceptual compression extracting the original data statistics
$\bm{\mu}$ and $\bm{\sigma}$ in a reduced dimension and use them as semantic vectors through a communication channel. 
Although this approach paved the way for the use of generative deep learning in SC, it has several drawbacks. The latent dimensionality cannot be dynamically changed according to network conditions since any further compression of data corresponds to a significant degradation in perceptual reconstruction performance. In addition, performance is strongly influenced by training data and the communication environment, making it effective only under certain predefined usage conditions. Lastly, VAEs often have limited expressive capacity, and therefore newer approaches use diffusion models to overcome these limitations.

Recently, a large number of new approaches that use generative models have been proposed in the field of semantic communication. Their application is tailored to almost any kind of multimedia content, especially for images \cite{Grassucci2023GenerativeSC,Han2022GenHighEff} and audio \cite{Grassucci2023DiffusionMF,Jang2024ICASSP}.
Interestingly, Hyelin Nam \textit{et al}. propose to use text as a semantic vector through a communication channel \cite{nam2024language}. They design a novel framework of language-oriented semantic communication. This innovative approach involves the use, at the sender side, of an image-to-text encoder that encapsulates the semantic meaning of an image into its corresponding textual caption. 
The receiver, in turn, is equipped with a huge text-to-image diffusion model that, taking the textual caption as input, generates an image with a similar semantic meaning with respect to the original one.
Even though their framework is quite efficient in terms of resources needed to exchange information (bandwidth and latency), the semantic gap between original and reconstructed data may be very large. Image-to-text models, for example, are not able to produce textual descriptions that capture all the details present in the original content. As a consequence, the reconstructed images may be vague, less detailed, and perceptually very dissimilar from the original ones.

To address the low perceptual similarities in language-oriented communications, in this paper, we introduce a novel framework communicating both (i) the textual caption and (ii) the latent embedding of the image. This is inspired by Stable Diffusion \cite{Rombach2021latent}, a latent diffusion model for text-to-image generation, which first generates a compressed image from (iii) pure noise in a latent space via iterative denoising while conditioning on (i), followed by scaling up to the original image resolution via an image decoder. In our proposed method, the denoising process starts not from (iii) but from (ii) that imposes high-fidelity semantic information such as spatial locations, color, and text, complementing textual caption's low-fidelity semantic information.

We validate our approach through experiments for the image-to-image communication task. We demonstrate that the embeddings of an image generated by an encoder network effectively retain the key features of the original image. For this reason, they are very useful to reconstruct data. On the receiver side, we adopt Stable Diffusion \cite{Rombach2021latent}, a well-known generative latent diffusion model that can be conditioned on text, image embedding, or both to generate more semantic accurate data. The conditioning choice can also be made dynamically based on network conditions. If the infrastructure is overloaded or the performance is very degraded we can send only a few characters describing original data. As soon as the network becomes better, we can send the latent embedding to boost the generative performance of the diffusion model. 

\begin{figure*}[!t]
\centering
\includegraphics[width=0.90\linewidth,keepaspectratio]{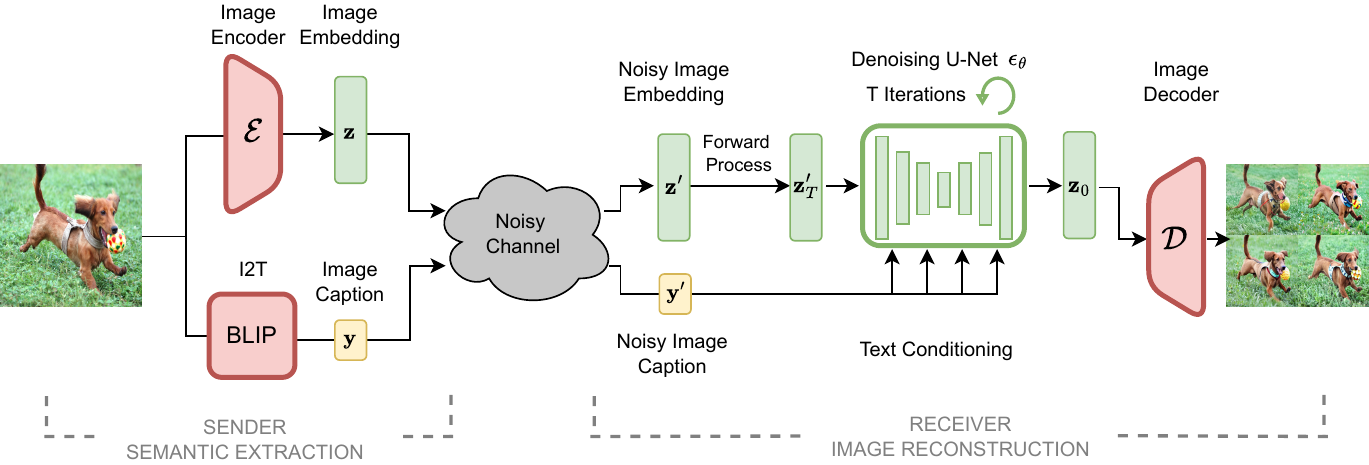}
\caption{Overview of the proposed framework. At the sender side, we employ both an image-to-text (I2T) model and an image encoder network. The I2T model produces the textual image caption while the image encoder encapsulates the scheme and the intrinsic semantics of an image in a latent representation. We transmit over the noisy network both text and image embeddings. At the receiver side, the content is regenerated using a latent diffusion model. This generative model takes noisy image embedding and applies $T$ diffusive steps conditioned on textual captions. At the end, the decoder network brings back the image to the original dimensionality. }
\label{fig_framework}
\vspace{-0.4cm}
\end{figure*}

\section{The Proposed Framework}
\label{sec:format}
In this Section, we accurately define the proposed semantic communication framework by addressing separately the two main components: semantic extraction at the sender side and data reconstruction at the receiver side.

\subsection{Semantic Latent Representation}
In our novel approach, we identify as semantic vectors both textual caption and latent embedding of an input image. This idea comes from the fact that, in most cases, the caption alone cannot fulfill the intrinsic semantic of an image and can lead to poor or vague representations of it. Consequently, our semantic encoder comprises two main components: an image-to-text (I2T) converter and an image encoder. 

\vspace{0.2cm}
\noindent\textbf{Image-to-text:} The I2T is responsible for translating an image $\textbf{x}$ into a text prompt $\textbf{y}$. The latter is defined as a sequence containing $Y$ words presented in a specific order:
\begin{equation}
    \textbf{y} = \text{I2T}(x) = (y_1,y_2, \dots, y_{Y})
\end{equation}
where $y_i$ is the $i$-th word and it has $|y_i|$ characters.
As image-to-text encoder, we use the well-known vision-language pre-training framework BLIP \cite{li2022blip}, which can produce accurate captions of a given image.

\vspace{0.2cm}
\noindent\textbf{Image encoder:} The image encoder $\mathcal{E}$ transforms an image $\textbf{x} \in \mathbb{R}^{C \times H \times W}$ from the RGB space into a latent representation $\mathbf{z} = \mathcal{E}(\mathbf{x})$. The literature is full of image encoders that can be employed. The choice is related to the domain of interest of the final applications and may be tailored for different kinds of tasks. In our study, we involve a pre-trained encoder from Stable-Diffusion v1.5 \cite{Rombach2021latent}.  Inspired by \cite{esser2021taming}, this encoder is part of an encoder-decoder network trained using a combination of a perceptual loss and a patch-based adversarial objective. The latent space learned from this type of model is of really small dimensionality compared to the initial RGB space. Despite this, experiments demonstrate that they are semantically and~perceptually equivalent \cite{esser2021taming}.

In our proposed framework, we consider transmitting text prompt $\textbf{y}$ along with latent image embedding $\textbf{z}$ over an additive white Gaussian noise channel (AWGN). The received $\textbf{y}$ and $\textbf{z}$ are independently distorted by a zero-mean Gaussian noise, and the channel conditions are identified by the signal-to-noise ratio (SNR) as $\text{SNR} = \frac{\textbf{P}_{signal}}{\textbf{P}_{noise}}$ where $\textbf{P}_{signal}$ and $\textbf{P}_{noise}$ are average received signal and noise powers, respectively. Note that the aggregate size of $\textbf{y}$ and $\textbf{z}$ is only $2.09$\% of the original image $\textbf{x}$ in our experiment, underscoring the bandwidth efficiency of our proposed method. Extending this to adaptive transmissions, it could be interesting to send only $\textbf{y}$ under extremely limited bandwidth or poor channel conditions, which is deferred to future studies.

\subsection{Data Reconstruction}
We equip the receiver with a conditional image generative model. We involve Stable Diffusion as it has demonstrated superior abilities in generating images from text. The methodology adopted by Stable Diffusion is based on a Latent Diffusion Model (LDM) \cite{Rombach2021latent}. It is a probabilistic model that can generate high-quality images starting from random noise and gradually transforming it into images. The key innovation of latent diffusion models is that they apply the diffusion process not to the raw pixel values but instead to an encoded latent representation of the image itself. This process can be conditioned on different signals (text, images, semantic maps, etc.). Commonly, LDMs comprise three modules: an encoder responsible for generating latent vectors, a U-Net devoted to the diffusion denoising process, and a decoder that brings latent vectors back to the image space. The novel idea behind our proposed framework is to distribute these modules between the sender and the receiver. The encoder is used in the sender to encapsulate the image in a latent representation. The U-Net and the decoder are used to regenerate content at the receiver side.

Following the theory of diffusion models, 
the conditioned U-Net $\epsilon_{\theta}(\textbf{z}_t,t,\textbf{y})$ has learned during training how to denoise the latent vector from pure Gaussian noise $\textbf{z}_T$ to $\textbf{z}_0$ in $T$ sampling steps. To reach this goal, a Markovian forward process is defined which injects noise at different time steps to the original latent vector according to:

\begin{equation}
        \mathbf{z}_t = \sqrt{\bar{\alpha_t}}\mathbf{z}_0 + \sqrt{1-\bar{\alpha_t}} \bm{\epsilon} \qquad \bm{\epsilon} \sim \mathcal{N}(\mathbf{0},\mathbf{I})
    \label{reparametrization}
\end{equation}
where $\{\bar{\alpha_t}\}_{t=1}^T$ is the noise schedule and controls the amount of injected noise.
The U-Net $\epsilon_{\theta}(\textbf{z}_t,t,\textbf{y})$ is trained to predict the amount of noise injected at any given time step using the objective function:

\begin{equation}
    \mathcal{L}_{DM}= \mathbb{E}_{z,y,\epsilon \sim \mathcal{N}(\textbf{0},\textbf{I}),t} \biggl[ ||\epsilon - \epsilon_\theta(\textbf{z}_t,t,\textbf{y})||_2^2\biggr]
\end{equation}

To accommodate the conditioning of text prompts, the U-Net is equipped with cross-attention mechanisms. Given text prompt $\textbf{y}$, CLIP \cite{radford2021learning} text encoder $\tau_\theta $ is used to produce its textual representation $\tau_\theta(\textbf{y})$ that is then injected into the cross-attention layers of the U-Net. In particular, key-value pairs (\textbf{K}, \textbf{V}) are built by projecting the text representation, while the query (\textbf{Q}) is built using the $i$-th intermediate representation of the U-Net $\epsilon_\theta$: 
\begin{equation}
\label{eq:crossattention}
    \Att(\textbf{Q}, \textbf{K}, \textbf{V}) = \softmax(\frac{\textbf{QK}^T}{\sqrt{d}})\times \textbf{V}
\end{equation}
\begin{equation}
    \textbf{Q}=\textbf{W}_Q\times \phi_i(\textbf{z}_t), \quad \textbf{K}=\textbf{W}_K\times \tau_\theta(\textbf{y}), \quad \textbf{V}= \textbf{W}_V\times \tau_\theta(\textbf{y})
\end{equation}
Here, $\phi_i(\textbf{z}_t)$ denotes the $i$-th intermediate representation of the U-Net implementing $\epsilon_\theta$. $\textbf{W}_Q,\textbf{W}_K,\textbf{W}_V$ are learnable matrices of parameters.

In our framework, the receiver acquires from the sender either a noisy version of the text prompt $\textbf{y}'$ or a combination of a noisy text prompt and noisy latent vector $(\textbf{y}',\textbf{z}')$. In the first case, the system starts the sampling phase from a random latent vector sampled from a normal Gaussian distribution $\textbf{z}_T \sim \mathcal{N}(\textbf{0}, \textbf{I})$. In the second case, it takes the vector $\textbf{z}'$, injects noise on it to obtain a noisy version $\textbf{z}'_T$ according to \eqref{reparametrization} and starts the generation from it.
There is a substantial and important difference between the two approaches. In the first case, the system needs to regenerate the content from pure noise conditioned on text prompt. In the second case we can control the amount of noise injected to $\textbf{z}'$ setting the number of generative sampling steps to any intermediate number $t$ from $1$ to $T$. Low sampling steps allow for an efficient generation (less time) but the final latent vector $\textbf{z}_0$ could retain the effects of the noise introduced by the network. Large sampling steps require more computational resources but allow for a more fine-grained generation.

After the latent diffusion process, a decoder model $\mathcal{D}$ takes the latent variable $\textbf{z}_0$ and upsample it. This module brings the latent variable back to the original RGB space $\hat{\textbf{x}}=\mathcal{D}(\textbf{z}_0)$,  $\hat{\textbf{x}} \in \mathbb{R}^{C \times H \times W}$.

\begin{figure}[!t]
\centering
\includegraphics[width=\linewidth]{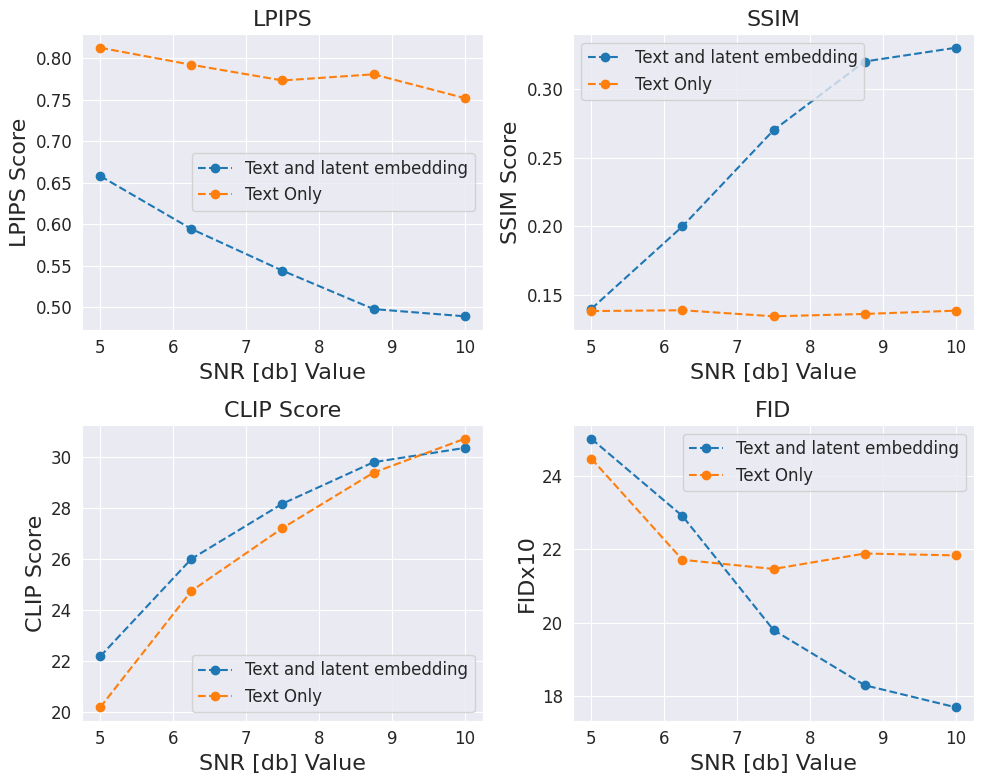}
\caption{Comparison between the two proposed approaches in a noisy channel scenario. Metrics taken into consideration: LPIPS, SSIM, CLIP Score and FID. For LPIPS and FID the lower the best. For SSIM and CLIP Score the higher the best.}
\label{fig_line}

\end{figure}

\section{Experimental results}

\subsection{Simulation Settings}  %In all the experiments we assume that only semantic information is exchanged between sender and receiver. Nothing from the originally intended images is conveyed, simulating a worst-case scenario in which a completely failed image transmission happens. 
We use BLIP-large \cite{li2022blip} as an I2T encoder. We leverage the power of the Stable Diffusion v1.5 image encoder at the sender side and denoising U-Net and decoder at the receiver side. We set $T=50$ denoising steps for text-only conditions while $T=30$ for text and latent embedding scenario.
The entire diffusion process is conditioned by the text prompt encoded using CLIP \cite{radford2021learning}. In order to validate our method we consider the Flickr 8k dataset \cite{Hodosh_Young_Hockenmaier_2013}. It contains 8,092 samples, each with a different shape. We rescale each image at 512x512 for convenience. For text prompts transmission, each character is 8-bit ASCII coded and modulated using 16QAM. For latent embedding transmission, each float number is 64-bit encoded as defined in the IEEE 754-2008 standard and modulated as well using 16QAM.
The metrics used to assess the quality of generated images are: Structural similarity index measure (SSIM), learned perceptual image patch similarity (LPIPS), Fréchet inception distance (FID), and CLIP score.
SSIM evaluates the distortion rate between $\textbf{x}$ and $\hat{\textbf{x}}$ at the pixel level. LPIPS and FID assess the perceptual similarity of intended and regenerated images using additional neural networks. CLIP score computes the correlation between texts and images.
%To assess the perceptual similarity between the sender intended image $\textbf{x}$ and the receiver generated one $\hat{\textbf{x}}$ we use the learned perceptual image patch similarity (LPIPS) score \cite{Zhang2018TheUE}. This metric calculates the distance at hidden layers of pre-trained AlexNet, given as:
%\begin{equation}
%    \operatorname{LPIPS}(\mathbf{x}, \hat{\mathbf{x}})=\sum_l \frac{1}{H_l W_l} \sum_{h, w} \| f(\mathbf{x}^l_{hw})-f\left(\hat{\mathbf{x}}^l_{hw}) \|_2^2 \right.
%\end{equation}
All the metrics are averaged over 100 simulation runs.

\begin{figure}[!t]
\centering
\includegraphics[width=\linewidth]{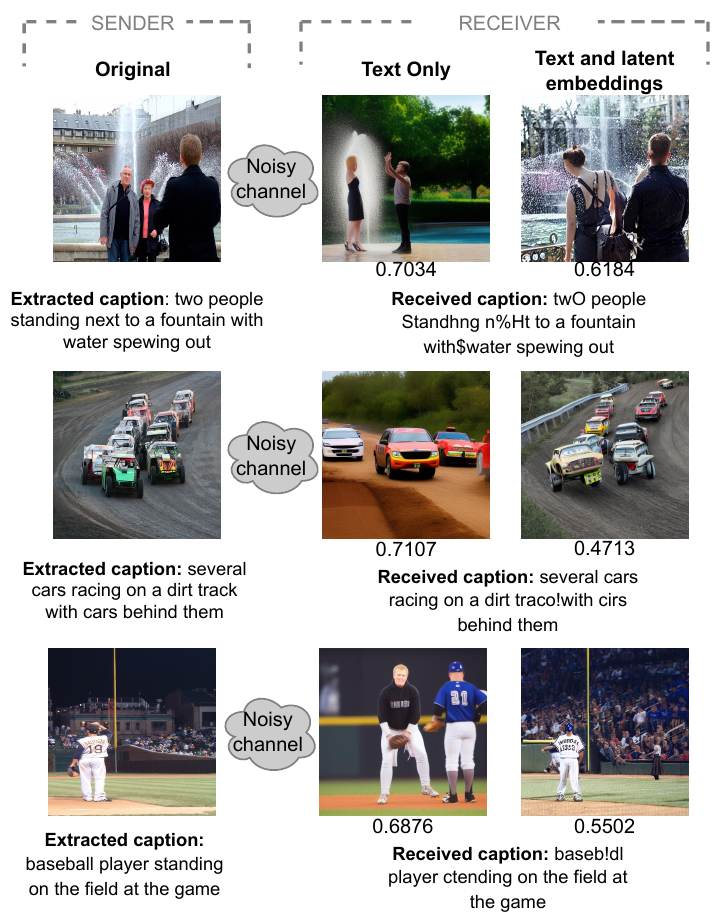}
\caption{Visual results. On the left-hand side three randomly selected samples along with text captions automatically extracted by BLIP-large model \cite{li2022blip}. On the right-hand side, there are reconstructed images. Numbers under images refer to the LPIPS score between regenerated images and intended ones. The SNR value is set to 7.5 dB.  }
\label{fig_comparison}
\end{figure}

\begin{figure*}[!t]
\centering
\includegraphics[width=0.85\linewidth,keepaspectratio]{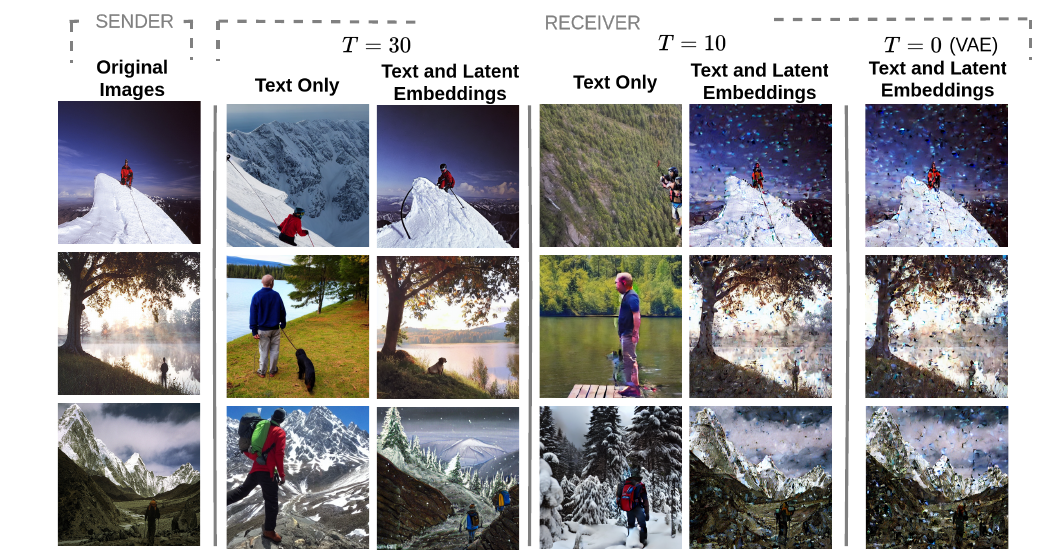}
\caption{ Visual results for different sampling timesteps. On the left-hand side three randomly selected samples. On the right-hand side reconstructed images using different strategies. Timesteps used are $T=[30,10,0]$. SNR value is set to 7.5 dB.   }
\label{fig_results_timesteps}
\vspace{-0.4cm}
\end{figure*}

\subsection{Experiments}

We begin our experiments by investigating the robustness of our method toward adverse channel conditions. As illustrated in Fig.~\ref{fig_line}, when there is a lot of noise introduced by the channel, both methods perform poorly. However, the joint text and latent embedding strategy seems to produce slightly better images.
On the contrary, when network conditions improve, our method clearly outperforms the text caption-only strategy. There is an improvement in SSIM  metric from $0.1387$ to $0.3321$ at $\text{SNR}=10 \text{ dB}$. In addition, there is a marked improvement in perceptual metrics LPIPS and FID leaving untouched the correlation between text and image (CLIP score).
Visual results can be appreciated in Fig.~\ref{fig_comparison}, where we display three random samples and we set $\text{SNR}=7.5 \text{ db} $.

We continue the experimental phase by comparing the performance of the two approaches using different sampling steps. In Fig.~\ref{fig_results_timesteps}, we compare the generation of three random images at different timesteps setting $\text{SNR}=7.5 \text{ dB}$.
It is interesting to see the role of the noise introduced by the channel. In the text and embedding scenario, if $T=0$,  the entire system collapses to a VAE 
acting as a transceiver, as no diffusion steps are performed. However, the noise introduced by the channel makes difficult the regeneration phase at the receiver side. On the contrary, when $T$ increases, the contribution of the diffusion model becomes clearer, proving its crucial role in eliminating this kind of noise. However, when $T$ grows too much, the generated images lose fidelity with respect to the original ones. This is because the cross-attention mechanism in \eqref{eq:crossattention} introduces more and more text conditioning during the regeneration process. Indeed, increased sampling steps correspond to heightened text conditioning and diminished latent embedding guidance, meaning that the generation is less faithful to the image embedding. For this reason, in our work, we involve $T=30$ sampling steps. 

%A judicious trade-off must be selected between excessive sampling steps, leading to less semantically aligned images, and insufficient sampling steps, resulting in the regeneration of noisy images.

We conclude our experimental investigation by analyzing the dimensionality of the data exchanged between sender and receiver.
Tab.~\ref{table_dimensionality} reveals that the text only 
is the most lightweight approach since at maximum we can exchange $77$ characters. This limitation comes from the $max\_tokens$ hyperparameter of CLIP text encoder. Sending the latent vector requires more computational resources (bandwidth and latency). However, despite the larger size, using the image embedding to regenerate the content allows us for a more accurate generation.  We can think of sending text and latent embeddings when the network conditions are good, while in case of bad conditions, we can send only the text, which retains much of the initial semantics.

\begin{table}[t]
\centering
\caption{Dimensionality comparisons.}
\resizebox{\linewidth}{!}{%
\begin{tabular}{c|ccc}
\hline
\multirow{2}{*}{\begin{tabular}[c]{@{}c@{}}Transmitted \\ Data\end{tabular}} & \multirow{2}{*}{Dimensionality} & \multirow{2}{*}{Data Type}  &\multirow{2}{*}{\begin{tabular}[c]{@{}c@{}}Payload Size\\ (Bytes)\end{tabular}}\\
 &  &   &\\ \hline
Whole image & {[}3,512,512{]} & Float 64  &   6.291.456 \\
Text Only & {[}77{]} & Int 8  &   616 \\
Latent embeddings & {[}4,64,64{]} & Float 64  & 131.072  \\
\begin{tabular}[c]{@{}c@{}}Text and \\ latent embeddings\end{tabular} & {[}4,64,64{]}+{[}77{]} & \begin{tabular}[c]{@{}c@{}}Int 8 and\\ Float 64\end{tabular}  & 131.688\\ \hline
\end{tabular}
}
\vspace{-0.5cm}
\label{table_dimensionality}
\end{table}

\vspace{-0.20cm}
\section{Conclusion}
In this paper, we propose an innovative framework for semantic image-to-image communication. We identify as a semantic vector to be conveyed in a noisy channel the textual caption along with a latent representation of input images. We leverage the generation power of an LDM to regenerate images semantically and perceptually aligned at the receiver side. Experiments validate our approach demonstrating a sensible reduction of the LPIPS metric when both text and latent embeddings are used to regenerate the content. Future research might improve the efficiency of our approach by reducing the dimensionality of latent embeddings and exploiting large language models (LLM) to compress the characters exchanged between sender and receiver, as suggested by \cite{nam2024language}. It could also be interesting to extend this work to other types of multimedia content such as audio, speech, and video.

\ninept
\bibliographystyle{IEEEbib}
\bibliography{biblio}

\end{document}